\documentclass[conference]{IEEEtran}
\IEEEoverridecommandlockouts
\usepackage{cite}
\usepackage{amsmath,amssymb,amsfonts, dsfont}
\usepackage{algorithm, algpseudocode, booktabs}
\usepackage{graphicx, subcaption}
\usepackage{float}
\usepackage{textcomp}
\usepackage{xcolor}
\usepackage{hyperref}
\usepackage{todonotes}

\def\BibTeX{{\rm B\kern-.05em{\sc i\kern-.025em b}\kern-.08em
    T\kern-.1667em\lower.7ex\hbox{E}\kern-.125emX}}
\begin{document}

\title{Deep Learning for Koopman Operator Estimation in Idealized Atmospheric Dynamics}

\author{\IEEEauthorblockN{David Millard}
\IEEEauthorblockA{\textit{Golisano College of CIS} \\
\textit{Rochester Institute of Technology}\\
Rochester, NY, USA \\
djm3622@rit.edu}
\and
\IEEEauthorblockN{Arielle Carr}
\IEEEauthorblockA{\textit{Computer Science \& Engineering} \\
\textit{Lehigh University}\\
Bethlehem, PA, USA \\
arg318@lehigh.edu}
\and
\IEEEauthorblockN{Stéphane Gaudreault}
\IEEEauthorblockA{\textit{Recherche en prévision numérique atmosphérique}\\
\textit{Environnement et Changement climatique Canada} \\
Dorval, Qc, Canada \\
stephane.gaudreault@ec.gc.ca}
}

\maketitle

\begin{abstract}
Deep learning is revolutionizing weather forecasting, with new data-driven models achieving accuracy on par with operational physical models for medium-term predictions. However, these models often lack interpretability, making their underlying dynamics difficult to understand and explain. This paper proposes methodologies to estimate the Koopman operator, providing a linear representation of complex nonlinear dynamics to enhance the transparency of data-driven models. Despite its potential, applying the Koopman operator to large-scale problems, such as atmospheric modeling, remains challenging. This study aims to identify the limitations of existing methods, refine these models to overcome various bottlenecks, and introduce novel convolutional neural network architectures that capture simplified dynamics.
\end{abstract}

\begin{IEEEkeywords}
dynamical systems, Koopman operator, autoencoders, convolutional neural networks
\end{IEEEkeywords}

\section{Introduction}

With the advancement of deep learning, numerous models have been developed to improve the efficiency of weather forecasting. Models like Pangu \cite{bi2022pangu},  Fourcastnet \cite{pathak2022fourcastnet}, and AIFS \cite{lang2024aifs} provide highly accurate medium-term predictions but come with the significant drawback of limited interpretability. The underlying dynamics of these models are often not well-understood, making them difficult to explain. Interpretability of data-driven models is essential for weather forecasting, especially in the context of decision-making in critical weather events. Also, as the importance of understanding climate change grows in the coming decades, it becomes increasingly crucial for scientists to grasp these dynamics.

In related studies, deep learning has been employed to discover Green's functions that solve for the time evolution of a dynamical system; see, e.g., \cite{bar2019learning, zhu2019physics}. A similar analogy could be made with the Koopman operator\cite{budivsic2012applied}, which offers a theoretical framework for linearizing dynamical systems. The Dynamic Mode Decomposition method \cite{doi:10.1137/1.9781611974508} is a prime example of an interpretable numerical approximation of the Koopman operator. The effectiveness of these approaches in various domains suggests their potential application to numerical weather forecasting, a field where interpretability is paramount.

Previous work by Lusch et al. \cite{Lusch2018} has demonstrated the feasibility of using deep learning and large datasets to estimate the Koopman operator. However, their approach encounters significant limitations when applied to large models and complex datasets, as commonly found in real-world weather prediction. Although convolutional neural networks (CNNs) have shown promise in this domain \cite{Xiao_2023}, they also present challenges, particularly in terms of information loss when handling complex tasks. This limitation results in networks that struggle to capture low-level dynamics, often leading to models that perform well on simpler tasks but fall short when faced with more complex challenges. We aim to bridge this gap and estimate Koopman operators for much more complicated tasks.

Our goals are twofold: to achieve predictive accuracy and to gain a deeper understanding of the underlying dynamics. To address both, we focus on estimating the Koopman operator and propose a novel CNN architecture capable of capturing simplified dynamics. As a step forward, we concentrate on estimating partial variables, specifically the temperature variable with its dependence on density removed, chosen for its richness of information. 

The remainder of the paper is organized as follows: Section~\ref{preliminary} presents the data-driven model, Section~\ref{experiments} presents the numerical tests,  Section~\ref{conclusion} provides a summary of the main conclusions, and Section~\ref{code} gives a reference to the open source repositories we used to generate data.

\section{Preliminaries}\label{preliminary}
\subsection{Dynamics}
Solving idealized flows with the non-hydrostatic Euler equations is often the first step toward the construction of a comprehensive numerical weather prediction model. In this study, a 2D version of the system is used. The first equation represents mass continuity, the second and third are the horizontal and vertical momentum equations, the fourth is the thermodynamic equation, and the last is the equation of state for an ideal gas. Numerical solution of this system will be used as a reference solution to train and to validate our data-driven model.
\begin{align}
\label{eq:continuity}    \dfrac{\partial \rho}{\partial t} + \dfrac{\partial \rho u^1}{\partial x^1} +
        \dfrac{\partial \rho u^3}{\partial x^3} &= 0 \\
\label{eq:mom1}    \dfrac{\partial \rho u^1}{\partial t} + \dfrac{\partial(\rho u^1u^1 + p)}{\partial x^1}
        + \dfrac{\partial \rho u^1u^3}{\partial x^3} &= 0 
\end{align}
\begin{align}
\label{eq:mom2}    \dfrac{\partial \rho u^3}{\partial t} + \dfrac{\partial\rho u^1u^3}{\partial x^1}
        + \dfrac{\partial (\rho u^3u^3 + p)}{\partial x^3} + \rho g &= 0 \\
\label{eq:thermo}    \dfrac{\partial \rho \theta}{\partial t} +
        \dfrac{\partial \rho u^1 \theta}{\partial x^1} +
        \dfrac{\partial \rho u^3 \theta}{\partial x^3} &= 0 \\
\label{eq:state} p_0 \left( \frac{R_d \, \rho \theta}{p_0} \right)^\gamma &= p
\end{align}
where \(\rho\) is the fluid density, \(p\) is the pressure, \(u^1\) is the horizontal velocity, \(u^3\) is the vertical velocity, \(\theta\) is the potential temperature, \(t\) is time, \(x^1\) is the horizontal spatial coordinate,  \(x^3\) is the vertical spatial coordinate, ($g \approx 9.81$ m/s$^2$) is the gravitational acceleration, $R_d$ is the specific gas constant for dry air, $\gamma$ is the specific heat ratio and reference pressure $p_0  = 100000$ Pa.

\subsection{Koopman Operator} 
Traditionally, the Euler equations are solved using appropriate numerical methods. Here we propose an alternative data-driven approach that approximates the Koopman operator. The Koopman operator is an infinite-dimensional linear operator that advances observables of a nonlinear dynamical system, in this case the density, winds and temperature, through time \cite[see Chapter 3]{doi:10.1137/1.9781611974508}. It provides a framework for analyzing nonlinear dynamics using linear techniques by lifting the system into a higher-dimensional function space where the evolution appears linear. The Koopman approach allows for the use of linear transformations to advance systems governed by nonlinear dynamics. Although the Koopman operator itself is infinite-dimensional, in practice, it is often approximated by finite-dimensional representations of its leading eigenfunctions, which capture the dominant dynamics of the system. This operator is particularly useful in systems where state advancement is composted of complex nonlinear transformations, such as those found in weather systems.

Let \(x_k\) be the current state, \(x_{k+1}\) be the state advanced one step in time, and \(\mathbf{F}_t\left(\cdot\right)\) be the nonlinear dynamics that advance the state in time,
\begin{equation}
    x_{k+1} = \mathbf{F}_t \left(x_k\right). \notag
\end{equation}
We then consider a nonlinear transformation \(g\left(\cdot\right)\) and the linear, but infinite dimensional, Koopman operator \(\mathcal{K}\) such that
\begin{equation}
    \mathcal{K}g\left(x_k\right) = g\left(\mathbf{F}_t \left( x_k\right)\right) = g\left(x_{k+1}\right),\notag
\end{equation}
where the Koopman operator applies a linear transformation to advance the state forward in time on the space given by \(g\left(\cdot\right)\).

To make the Koopman operator practically useful, various methods have been developed to approximate it in finite dimensions. Dynamic Mode Decomposition \cite{doi:10.1137/1.9781611974508} identifies the leading modes of the system that dominate its behavior. Autoencoders can be used to estimate both \(g\left(\cdot\right)\) and \(\mathbf{K}^m\) with specific constraints. Through such approximations, the Koopman operator can be applied to a wide range of complex dynamical systems, offering insights into their underlying structure and facilitating predictive modeling.

\subsection{Autoencoders} 
An autoencoder is a neural network architecture consisting of three core components: an encoder, a latent space, and a decoder. The encoder transforms the input data into a nonlinear subspace, typically resulting in significantly reduced dimensionality. The latent space represents this transformed knowledge, where constraints can be added to enforce specific properties. The decoder then attempts to reverse the transformation, reconstructing the data from its latent space representation back into the original state space. Autoencoders are commonly used for tasks such as dimensionality reduction \cite{hinton2006reducing}, data denoising \cite{vincent2008extracting}, and generative modeling \cite{kingma2014auto}. The quality of the reconstruction relies on the capacity of the latent space and the ability of the encoder-decoder pair to capture the complex nonlinear relationships within the data.

We focus on an autoencoder structure proposed in \cite{Lusch2018} that enforces the latent space to be Koopman coordinates. We achieve this by constraining the encoder-decoder subnetwork to reconstruct the input data and the encoder-latent-decoder network to predict future states, where the latent space is a linear transformation. This approach allows the model to effectively estimate the nonlinear transformation to the Koopman coordinate space along with the Koopman operator that advances these alternate representations in time linearly. 

\begin{figure*}[t!]
    \centering
    \begin{minipage}{0.48\textwidth}
        \centering
        \includegraphics[width=\textwidth]{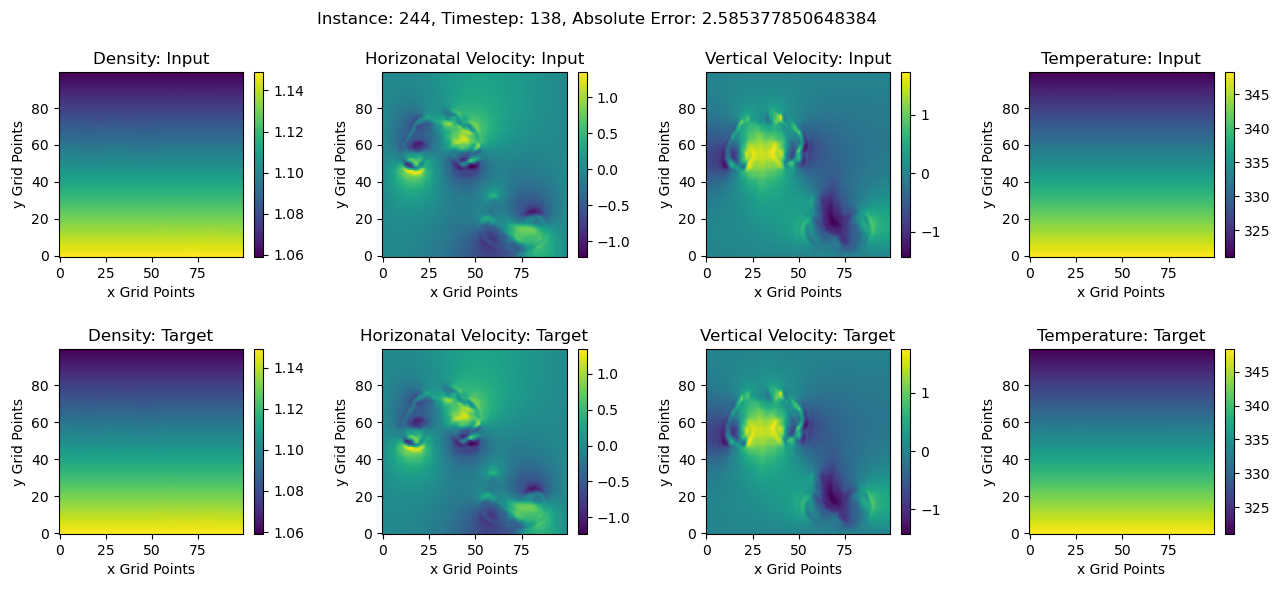}
    \end{minipage}\hfill
    \begin{minipage}{0.48\textwidth}
        \centering
        \includegraphics[width=\textwidth]{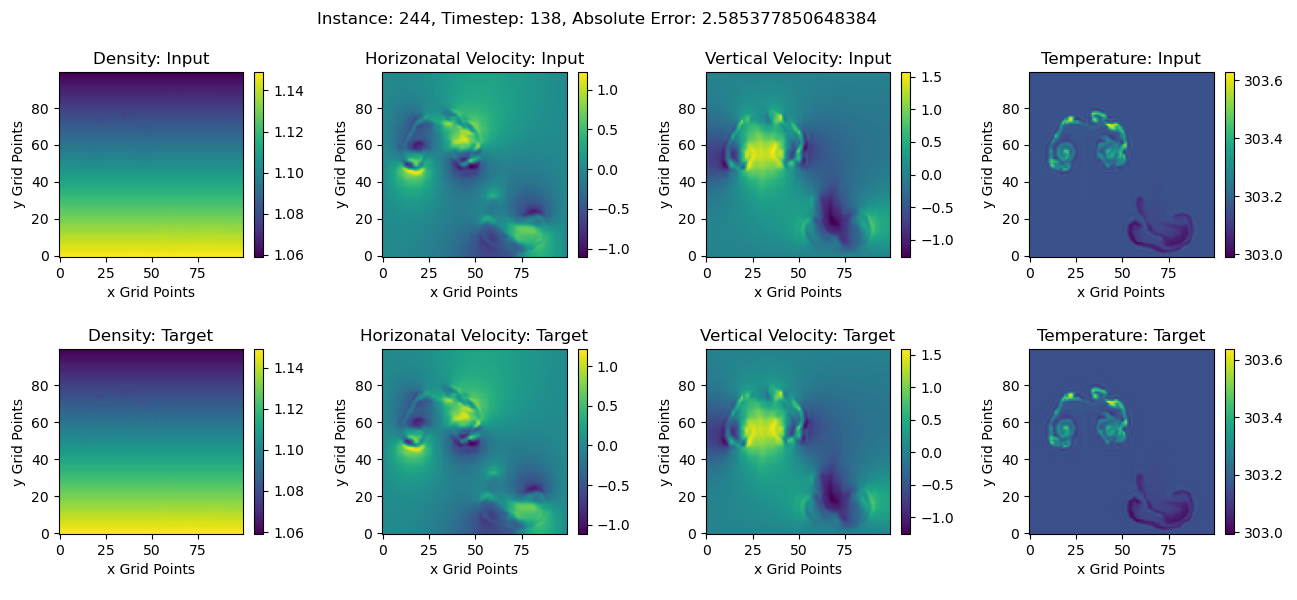}
    \end{minipage}
    \caption{On the left are the un-transformed variables density, horizontal velocity, vertical velocity, and temperature, respectively. On the right are the same four variables but transformed, except density. To transform the variables, the density values are divided out of each respective variable. It is important to note that the density variable is particularly significant. The density variable defines the structure of the bubble, while the temperature variable describes the heat distribution within that structure. Thus, the temperature variable reflects the heatmap of the system, with the density providing the underlying form.}
    \label{fig:data-full}
\end{figure*}

\subsection{Loss Functions}
As proposed in \cite{Lusch2018}, three losses are introduced to effectively train an autoencoder that estimates the nonlinear transformations alongside the Koopman operator. These constraints are defined as follows
\begin{align}
    \mathcal{L}_{\text{recon}} &= \left\Vert x_k -  g^{-1}\left(g \left( x_k\right)\right)\right\Vert_{\text{MSE}} \\
    \mathcal{L}_{\text{pred}} &= \left\Vert x_{k+1} -  g^{-1}\left(\mathbf{K}^m g\left( x_k\right)\right)\right\Vert_{\text{MSE}} \\
    \mathcal{L}_{\text{lin}}  &= \left\Vert g\left(x_{k+1}\right) -  \mathbf{K}^m g\left(x_{k}\right)\right\Vert_{\text{MSE}},
\end{align}
where \(x_k\) represents the input state vector, \(g\) denotes the encoder, \(g^{-1}\) is the decoder, \(\mathbf{K}^m\) represents the Koopman operator applied in the latent space, and MSE denotes the mean squared error. 
\(\mathcal{L}_{\text{recon}}\) minimizes the loss between the input state and the output of the encoder-decoder subnetwork, referred to as the reconstruction. \(\mathcal{L}_{\text{pred}}\) minimizes the loss between the input state and the encoder-latent-decoder output, referred to as the prediction. \(\mathcal{L}_{\text{lin}}\) minimizes the loss between the encoded advanced state and the encoder-latent state, constraining the model to be linear in the latent space.

In addition to the original constraints, we propose two supplementary losses that enhance generalization and perform regularization
\begin{align}
    \mathcal{L}_{\text{noise}} &= \left\Vert g\left(\Tilde{x}_{k+1}\right) -  \mathbf{K}^m g \left( x_k\right)\right\Vert_{\text{MSE}} \\
    \mathcal{L}_{\text{repl}} &= \left\Vert g\left(\Tilde{x}_{k+1}\right) -  g \left( x_{k+1}\right)\right\Vert_{\text{MSE}},
\end{align}
where \(\Tilde{x}_{k+1}\) represents a predicted next state. The \(\mathcal{L}_{\text{noise}}\) loss ensures that the encoded representations of advanced states remain consistent with their predicted encodings, thereby promoting stability in the latent space. The \(\mathcal{L}_{\text{repl}}\) loss aids in encoding noisy or slightly different states into a similar latent space, encouraging robustness to variations. These constraints are essential in cases involving high-complexity dynamics as the model may struggle to fully capture the intricate behavior of the data, leading to outputs not representative of the training data. Without these additional constraints, the model would be unable to encode the simplified dynamics into the same latent space it encountered during training.

Finally, the overall objective function is defined as a weighted combination of the loss terms
\begin{align} \label{eq:obj}
    \mathcal{L} = 
    a_1\mathcal{L}_{\text{recon}} + a_2\mathcal{L}_{\text{pred}} + a_3\mathcal{L}_{\text{lin}} + a_4\mathcal{L}_{\text{noise}} + a_5\mathcal{L}_{\text{repl}},
\end{align}
where \(a_1, a_2, a_3, a_4,\) and \(a_5\) are scalar multipliers that weight the contributions of each loss term. This objective aims to train the  autoencoder to learn the representation of the nonlinear dynamics along with an estimate of the Koopman operator.

\section{Experiments}\label{experiments}
\subsection{Test case}

Robert \cite{robert1993bubble} describes a theoretical model where euther a warm or a cold bubble evolves in a two-dimensional, dry isentropic atmosphere. To instantiate a bubble, we apply a Gaussian perturbation to the state. The state is then discretized and advanced through time. A timestep of 5 seconds is used with a grid of \(25 \times 25\) elements. A 4\textsuperscript{th}-order nodal discontinuous Galerkin spatial discretization \cite{hesthaven2007nodal} is used with a second-order Rosenbrock time integration scheme \cite{dallerit2024second}. This configuration of the WxFactory model \cite{GAUDREAULT2022110792} is utilized to generate both the training and the validation datasets.

\subsection{Data}
The data used consists of state vectors collected from a modified version of the model. In this modified version, the colliding bubbles configuration was set to randomly spawn one or two hot bubbles and between zero and two cold bubbles, each with a randomized radius. The bubbles were then assigned randomized temperatures within their respective limits for hot and cold, along with a randomized location—hot bubbles were typically set to spawn lower, while cold bubbles typically spawned higher. All parameter values/domains can be found in Table \ref{table:params}.

\begin{table}[h]
\centering
\begin{tabular}{|c|c|c|}
\hline
\textbf{Parameters} & \textbf{Hot Bubbles} & \textbf{Cold Bubbles}\\
\hline
State (\(m \times m \)) & \(\left(1000,1000\right)\) & \(\left(1000,1000\right)\)\\
\hline
Objects & \(o_h \in \left(1, 2\right)\) & \(o_c \in \left(0, 1, 2\right)\)\\
\hline
Temp. \((K)\) & \(k_h \in \left(303.3 \cdots 303.6 \right)\) & \(k_c \in \left(302.8 \cdots 302.9\right)\)\\
\hline
Radius \((m)\)& \(r_x \in \left(10 \cdots 80\right)\) & \(r_x \in \left(10 \cdots 80\right)\)\\
\hline
Stability &  \(50\) & \(50\)\\
\hline
\(x\)-Center \((m)\) & \(c_x \in \left(300 \cdots 700\right)\) & \(c_x \in \left(200 \cdots 800\right)\) \\
\hline
\(z\)-Center \((m)\) & \(c_z \in \left(50 \cdots 300\right)\) & \(c_z \in \left(100 \cdots 750\right)\) \\
\hline
\end{tabular}
\caption{Initialization parameter domains.}
\label{table:params}
\end{table}

After initialization, the states are advanced for 215 timesteps with the state vector saved to disk after each timestep. During generation, some timesteps encountered instability, preventing the computation of the subsequent step. In such cases, the number of generated timesteps minus ten was used. The state vectors are comprised of four different discretized variables: density (\(\rho\)), horizontal velocity (\(u^1\)), vertical velocity (\(u^3\)), and temperature (\(\theta\)). The state space consists of 10,000 solution points per variable arranged in a \(100 \times 100\) grid. Therefore, the total state space is structured as \(4 \times 100 \times 100\). Since each variable has a dependence on density, to obtain an accurate estimate of a single variable, \(\rho\) must be factored out from equations \eqref{eq:continuity}-\eqref{eq:thermo}.

The dataset is divided into training and validation sets. The validation set consists of 240 unique instances, while the training set contains 700 unique instances, each with up to 215 timesteps and with the median number of timesteps being this maximum. This configuration provides the model with approximately 200,000 total data points. To augment the training data while preserving the underlying physics, a random horizontal flip is applied within the data augmentation pipeline. Figure \ref{fig:data-full} dissects a complete instance.

Before the data is fed into the model, it is first pre-processed to normalize each variable, ensuring that all input values are the same orders of magnitude. The means and standard deviations used for normalization are saved for use during reconstruction and prediction.

\subsection{Experimental Setup}
The experiments were conducted using a system equipped with a single NVIDIA RTX A5000 GPU. The environment was configured with Linux, running Python 3.9.7 along with PyTorch 2.4.0. The model architectures were trained on a dataset consisting of approximately 150,000 state vectors. The training and validation split followed an 80:20 ratio, enabling a comprehensive assessment of the model's generalization capability. Early stopping was employed to determine when the model had converged. The terms ``pseudo-convergence" and ``pseudo-minimum" are used to describe instances when the model had not fully converged or reached the true minimum, but was no longer improving at a sufficient rate. The training process utilized the Adam optimizer with a constant learning rate of \(1 \times 10^{-8}\).

To evaluate the model's performance Mean Squared Error (MSE), the objective defined in equation (\ref{eq:obj}) and 2-norm were used. Equation (\ref{eq:obj}) is instantiated with scalar multipliers $a_{1},a_{2}, a_{3}, a_{4}$, and $a_{5}$ all equal to 1 and kept constant throughout the entire training process. In each of the following models, the estimator of $\mathbf{K}^m$ is denoted as a linear layer with no bias parameters.

\subsection{Related Work}
\begin{figure}[h]
    \centering
    \includegraphics[width=1.0\linewidth]{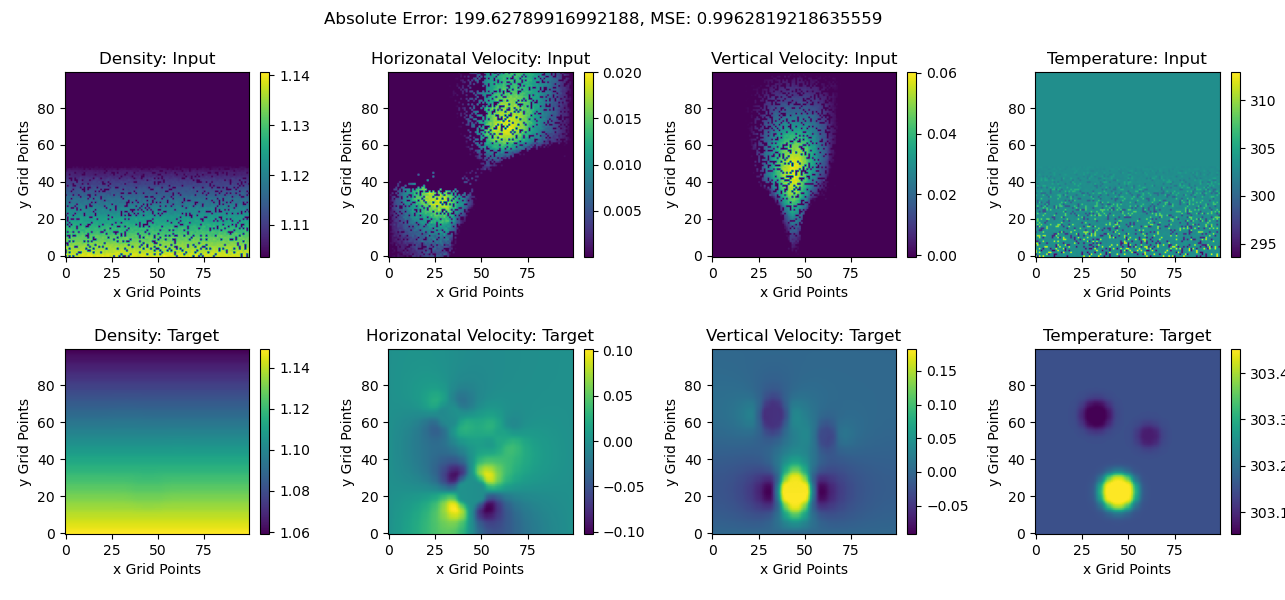}
    \caption{Reconstructions from a preliminary fully-connected model, inspired by \cite{Lusch2018}. The model is particularly limited by rapid dimensionality reduction when dealing with high-dimensional input data. It also fails to capture meaningful dynamics, resulting in crude estimates.}
    \label{fig:fullconn_out_bad}
\end{figure}

\begin{figure}[h]
    \centering
    \includegraphics[width=1.0\linewidth]{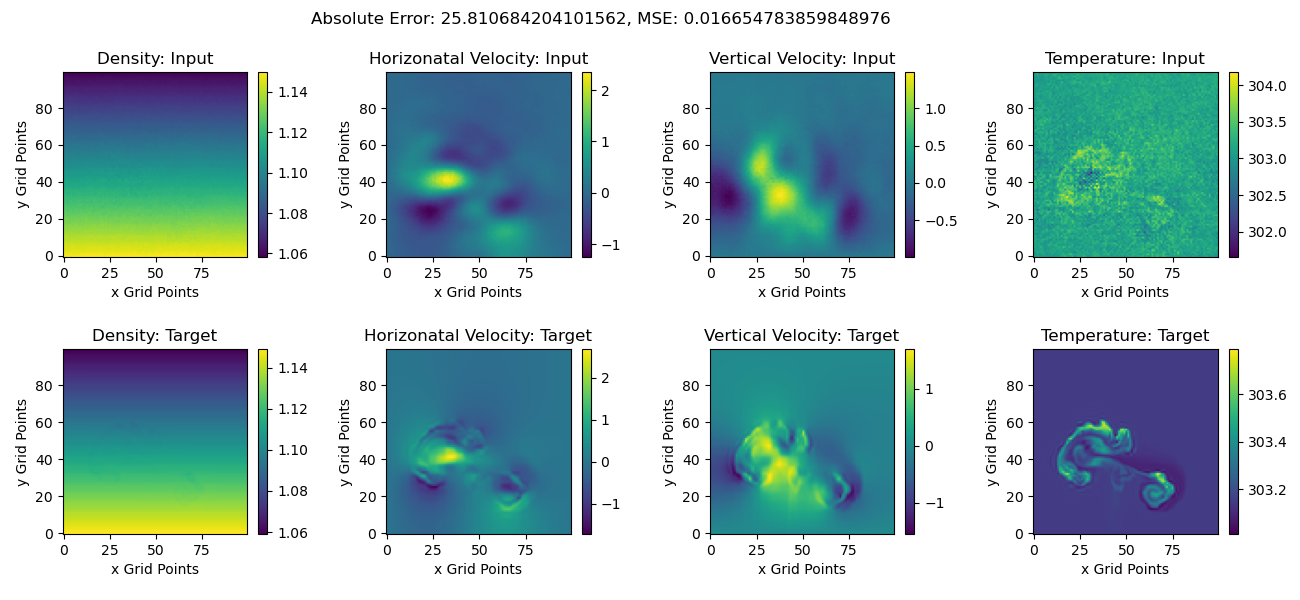}
    \caption{Reconstructions from a preliminary full-CNN model, inspired by \cite{Xiao_2023}.  The model is hindered by the complexity of the data, resulting in significant information loss. While it achieves a rough approximation of the data, it encounters challenges with the density and temperature variables.}
    \label{fig:cnn_full_out_bad}
\end{figure}

Lusch et al. \cite{Lusch2018} provide a foundational framework for estimating the Koopman operator; however, this approach faces significant challenges when applied to high-dimensional, complex data. In prior experiments, we explored a fully-connected linear model but found it inadequate due to hardware limitations when handling data of this scale; see Figure \ref{fig:fullconn_out_bad}. Yongqian Xiao et al. \cite{Xiao_2023} propose a novel CNN architecture similar to our approach, but we observed significant bottlenecks in the quality of reconstructions and predictions; see Figure \ref{fig:cnn_full_out_bad}. We discovered that estimating a single variable at a time -- provided the variable is sufficiently rich in information -- results in much better generalization of models. Though, this still does not fully solve the problem of advancing the complete system state.

\subsection{Partial-CNN AE}
\begin{figure*}[h]
    \centering
    \includegraphics[width=1.0\linewidth]{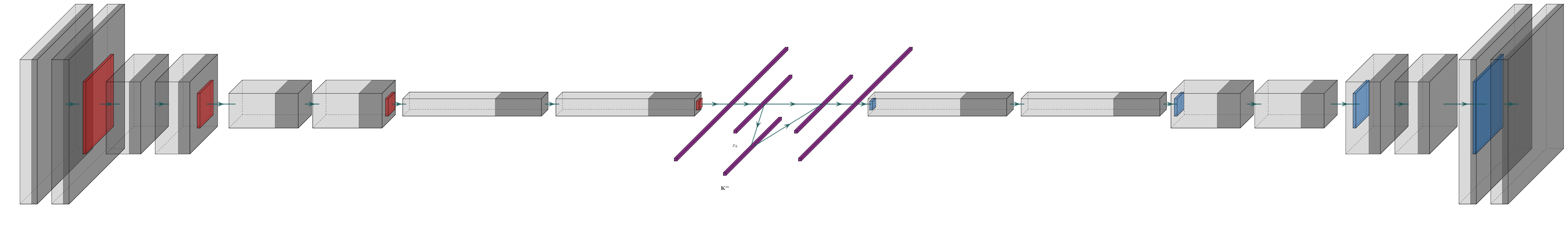}
    \caption{The novel CNN-AE architecture. On the left is the encoder, which consists of transparent blocks representing residual blocks and red blocks denoting max pooling layers. In the middle is the latent space. To reconstruct the input, the output from the first latent space bypasses \(\mathbf{K}^m\) and is directly passed to the subsequent latent space. To advance the state, the output from the first latent space is passed through \(\mathbf{K}^m\) before being fed into the next latent space. On the right is the decoder, which mirrors the encoder with transparent blocks representing residual blocks and blue blocks indicating transposed convolutions.}
    \label{fig:prof_cnn_full}
\end{figure*}

The partial-CNN autoencoder is described in Table \ref{table:cnnpart} and depicted in Figure \ref{fig:prof_cnn_full}.
A DownBlock consists of a sequence of two Residual Blocks \cite{he2015deepresiduallearningimage} followed by Max Pooling. An UpBlock comprises a Transposed Convolution followed by two Residual Blocks . The kernel size for each convolution is \(3 \times 3\) with a padding of 1 and a stride of 1. The pooling operation reduces the image size by a factor of 2. The Transposed Convolution employs a kernel size of \(2 \times 2\) with a stride of 2. The number of filters, image size after every block, and latent sizes are detailed in Table \ref{table:cnnpart}. The output from Latent 1 is denoted as \(x_k\). The advanced state \(x_{k+1}\) is obtained by passing \(x_k\) through the Latent \(\mathbf{K}^m\) layer, followed by the Latent 2 layer.

\begin{table}[h!]
\centering
\begin{tabular}{|c|c|}
\hline
\textbf{Layer}                     & \textbf{Size}              \\ \hline
Input         & $1 \times 100 \times 100$ \\ \hline
DownBlock 1         & $64 \times 50 \times 50$ \\ \hline
DownBlock 2         & $128 \times 25 \times 25$ \\ \hline
DownBlock 3        & $256 \times 12 \times 12$ \\ \hline
DownBlock 4        & $512 \times 6 \times 6$ \\ \hline
Flatten (Encoder Output)           & $18432$                   \\ \hline
Linear (Latent 1)                   & $18432 \times 4096$       \\ \hline
Linear ($\mathbf{K}^m$)                   & $4096 \times 4096$       \\ \hline
Linear (Latent 2)                & $4096 \times 18432$        \\ \hline
Reshape (Decoder Input)         & $512 \times 6 \times 6$ \\ \hline
UpBlock 4        & $256 \times 12 \times 12$ \\ \hline
UpBlock 3        & $128 \times 25 \times 25$ \\ \hline
UpBlock 2         & $64 \times 50 \times 50$ \\ \hline
UpBlock 1         & $4 \times 100 \times 100$ \\ \hline
Output    & $ 1 \times 100 \times 100$ \\ \hline
\end{tabular}
\caption{Layer sizes in partial-CNN autoencoder.}
\label{table:cnnpart}
\end{table}

\begin{figure}[h]
    \centering
    \includegraphics[width=1.0\linewidth]{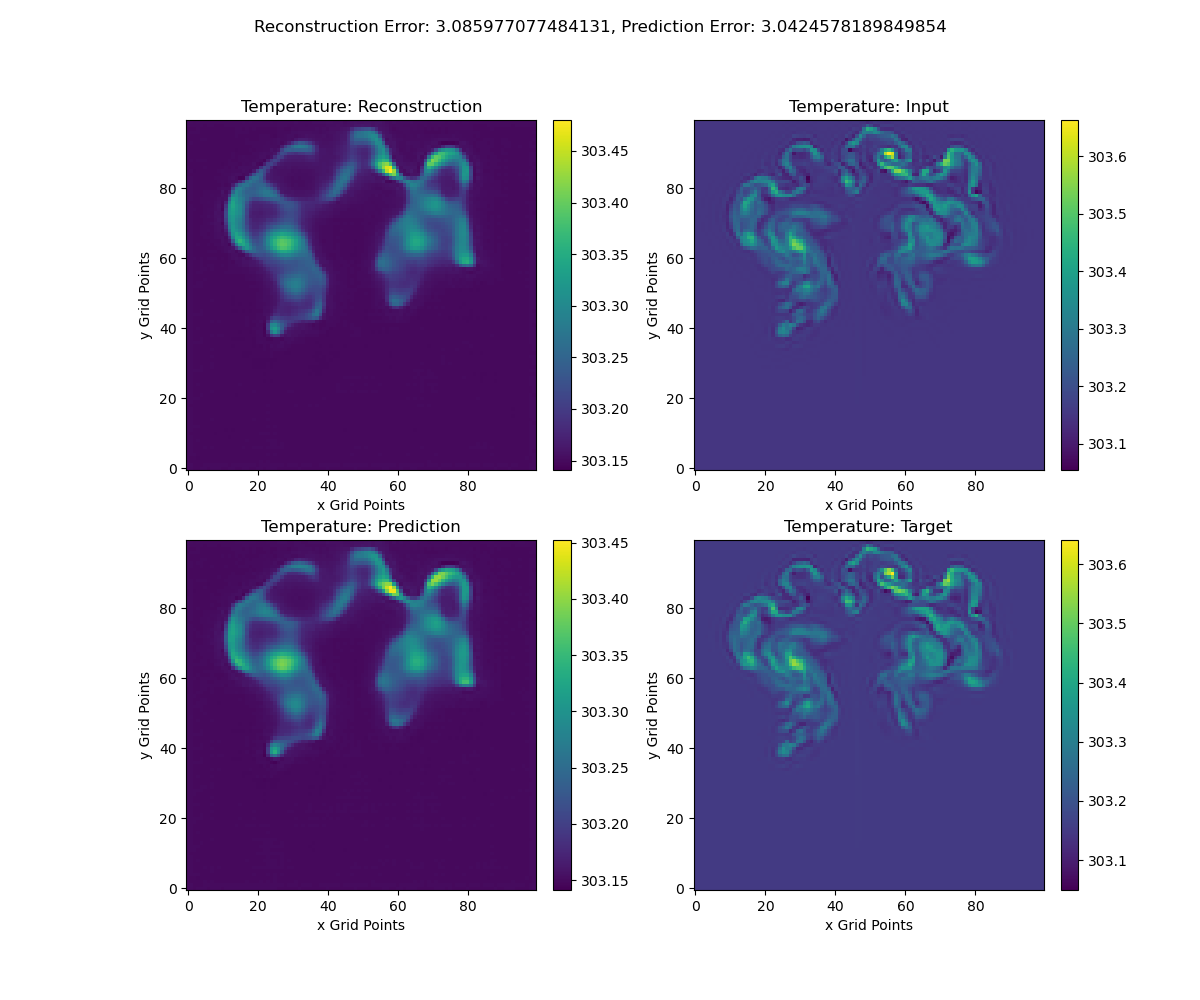}
    \caption{Partial-CNN autoencoder reconstruction and prediction of an instance from the validation set.}
    \label{fig:full_output}
\end{figure}

\begin{figure*}[]
    \centering
    \begin{minipage}{0.48\textwidth}
        \centering
        \includegraphics[width=\textwidth]{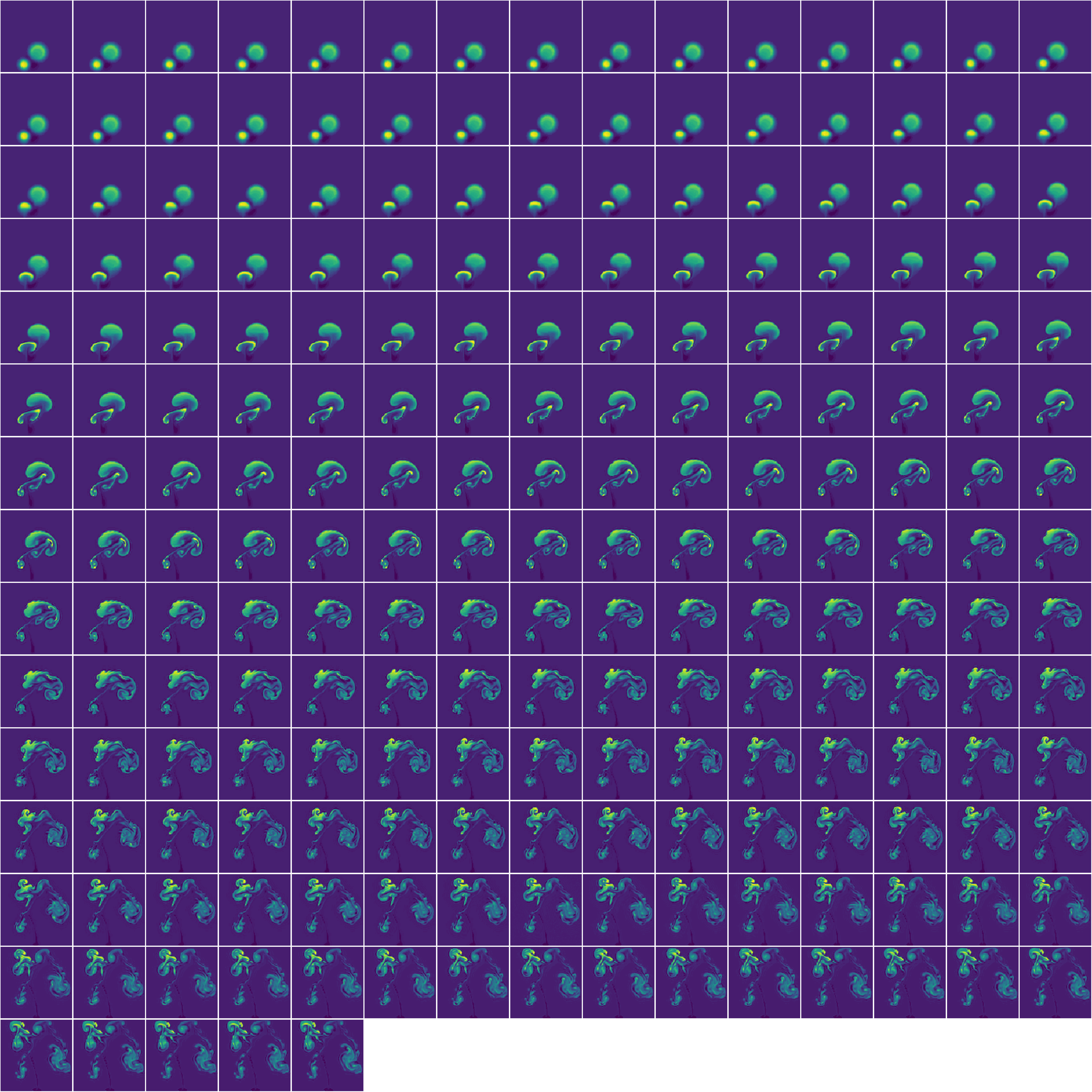}
    \end{minipage}\hfill
    \begin{minipage}{0.48\textwidth}
        \centering
        \includegraphics[width=\textwidth]{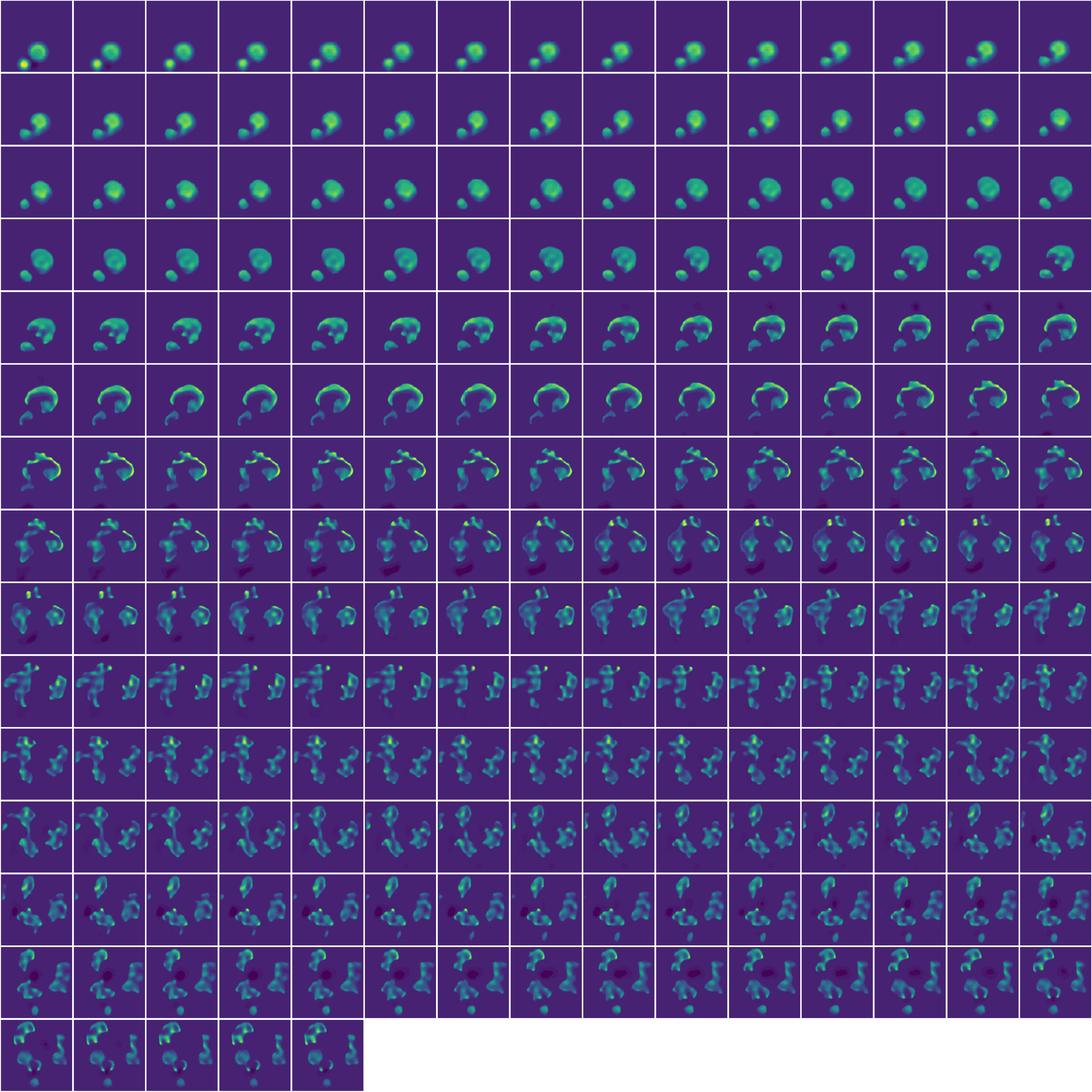}
    \end{minipage}
    \caption{Left: The ground truth of an initial state (top-left) advanced 215 timesteps. Right: A single-shot prediction using the Partial-CNN given the same initial conditions. The states are generated by taking each decoded state advancement from the autoencoder and feeding it back into the model. After each step, the state is saved.}
    \label{fig:stateadv}
\end{figure*}

The input to this model is exclusively the transformed temperature variable. The training process spanned approximately five hours, achieving pseudo-convergence at the 89th epoch. The model attained a pseudo-minimum objective loss, as described in equation \ref{eq:obj}, of approximately 0.25 on the validation set and 0.26 on the training set. Additionally, the lowest MSE loss recorded on the validation set was around 0.007. Figure \ref{fig:full_output} illustrates a reconstruction from the validation set with a 2-norm error of about 3.0. Figure \ref{fig:stateadv} provides a one-shot state advancement of using an initial transformed temperature variable along with its ground truth. These metrics and the results presented in Figure \ref{fig:full_output} indicate significant generalization. Despite utilizing only a single variable, which limits the model's ability to capture the full dynamics, it successfully captures most of the high- and medium-level features. Moreover, even with limited knowledge of the dynamics, the model can perform single-shot predictions using only an initial state, as shown in Figure \ref{fig:stateadv}. This approach, relying on a single state prediction, means the model has no information about the velocity component to assist in its estimation.

\subsection{Discussion}
Qualitative and quantitative evidence suggests that the partial-CNN model is capable of capturing localized simplified dynamics, as indicated by the rough predictions and the state advancement that diverges. This model is able to account for a portion of the highly complex nonlinear dynamics and estimate an operator that evolves the state. However, we acknowledge that this model does not solve the problem of estimating the entire state vector. Previous experiments revealed that attempting to estimate the full state space results in poor reconstructions. These models are quite large and fitting them to datasets with reduced diversity tends to impede the learning of key components such as convolution layers, which struggle in environments with limited data variation.

\section{Conclusion and Future Work}\label{conclusion}
In this paper, we present an approach for estimating the Koopman operator within the context of real-world weather data. While foundational methods, such as those proposed by Lusch et al. \cite{Lusch2018} and Xiao et al. \cite{Xiao_2023}, provide excellent frameworks for estimating dynamical systems, they encounter significant bottlenecks when faced with the complexity and scale of weather data. We demonstrate substantial progress in estimating the Koopman operator using only partial information and show its ability to capture short-term state advancements. Although the model's performance is limited, it serves as a pivotal stepping stone for future research. Specifically, it highlights the challenge of training models on data of this scale, with a particular focus on capturing low-level dynamics and long-term state advancements. Our findings indicate that while current methods offer partial solutions, a considerable gap remains in developing models that fully capture the underlying dynamics of complex weather systems.

The field of generative deep learning has long faced the challenge of accurately reconstructing and predicting complex, high-dimensional data. Although models like variational autoencoders, generative adversarial networks, and diffusion networks have potential in theory, effective training has been difficult. Recent advancements in open-source models and transfer learning have made it possible to apply these models to various tasks. Given the parallels between image generation processes and solving for the Koopman operator, transfer learning using these models could be highly effective in this context. We plan to pursue this line of research in future work. Techniques based on operator learning, in particular the Fourier neural operators \cite{li2020fourier,kovachki2023neural}, will also be investigated.

\section{Code Availability}\label{code}
The code of the WxFactory model is available under the Apache License version 2.0 at
\url{https://github.com/Wx-Alliance-Alliance-Meteo/WxFactory}.


\bibliographystyle{plain}
\bibliography{refs}

\begin{thebibliography}{10}

\bibitem{bar2019learning}
Yohai Bar-Sinai, Stephan Hoyer, Jason Hickey, and Michael~P Brenner.
\newblock Learning data-driven discretizations for partial differential equations.
\newblock {\em Proceedings of the National Academy of Sciences}, 116(31):15344--15349, 2019.

\bibitem{bi2022pangu}
Kaifeng Bi, Lingxi Xie, Hengheng Zhang, Xin Chen, Xiaotao Gu, and Qi~Tian.
\newblock Pangu-weather: A 3d high-resolution model for fast and accurate global weather forecast.
\newblock {\em arXiv preprint arXiv:2211.02556}, 2022.

\bibitem{budivsic2012applied}
Marko Budi{\v{s}}i{\'c}, Ryan Mohr, and Igor Mezi{\'c}.
\newblock Applied koopmanism.
\newblock {\em Chaos: An Interdisciplinary Journal of Nonlinear Science}, 22(4), 2012.

\bibitem{dallerit2024second}
Valentin Dallerit, Tommaso Buvoli, Mayya Tokman, and St{\'e}phane Gaudreault.
\newblock Second-order rosenbrock-exponential (rosexp) methods for partitioned differential equations.
\newblock {\em Numerical Algorithms}, 96(3):1143--1161, 2024.

\bibitem{GAUDREAULT2022110792}
Stéphane Gaudreault, Martin Charron, Valentin Dallerit, and Mayya Tokman.
\newblock High-order numerical solutions to the shallow-water equations on the rotated cubed-sphere grid.
\newblock {\em Journal of Computational Physics}, 449:110792, 2022.

\bibitem{he2015deepresiduallearningimage}
Kaiming He, Xiangyu Zhang, Shaoqing Ren, and Jian Sun.
\newblock Deep residual learning for image recognition, 2015.

\bibitem{hesthaven2007nodal}
Jan~S Hesthaven and Tim Warburton.
\newblock {\em Nodal discontinuous Galerkin methods: algorithms, analysis, and applications}.
\newblock Springer Science \& Business Media, 2007.

\bibitem{hinton2006reducing}
Geoffrey~E Hinton and Ruslan~R Salakhutdinov.
\newblock Reducing the dimensionality of data with neural networks.
\newblock {\em science}, 313(5786):504--507, 2006.

\bibitem{kingma2014auto}
Diederik~P Kingma and Max Welling.
\newblock Auto-encoding variational bayes.
\newblock In {\em 2nd International Conference on Learning Representations (ICLR)}, 2014.

\bibitem{kovachki2023neural}
Nikola Kovachki, Zongyi Li, Burigede Liu, Kamyar Azizzadenesheli, Kaushik Bhattacharya, Andrew Stuart, and Anima Anandkumar.
\newblock Neural operator: Learning maps between function spaces with applications to pdes.
\newblock {\em Journal of Machine Learning Research}, 24(89):1--97, 2023.

\bibitem{doi:10.1137/1.9781611974508}
J.~Nathan Kutz, Steven~L. Brunton, Bingni~W. Brunton, and Joshua~L. Proctor.
\newblock {\em Dynamic mode decomposition: Data-driven Modeling of Complex Systems}.
\newblock Society for Industrial and Applied Mathematics, Philadelphia, PA, 2016.

\bibitem{lang2024aifs}
Simon Lang, Mihai Alexe, Matthew Chantry, Jesper Dramsch, Florian Pinault, Baudouin Raoult, Mariana~CA Clare, Christian Lessig, Michael Maier-Gerber, Linus Magnusson, et~al.
\newblock Aifs-ecmwf's data-driven forecasting system.
\newblock {\em arXiv preprint arXiv:2406.01465}, 2024.

\bibitem{li2020fourier}
Zongyi Li, Nikola Kovachki, Kamyar Azizzadenesheli, Burigede Liu, Kaushik Bhattacharya, Andrew Stuart, and Anima Anandkumar.
\newblock Fourier neural operator for parametric partial differential equations.
\newblock {\em arXiv preprint arXiv:2010.08895}, 2020.

\bibitem{Lusch2018}
Bethany Lusch, J.~Nathan Kutz, and Steven~L. Brunton.
\newblock Deep learning for universal linear embeddings of nonlinear dynamics.
\newblock {\em Nature Communications}, 9(1):4950, Nov 2018.

\bibitem{pathak2022fourcastnet}
Jaideep Pathak, Shashank Subramanian, Peter Harrington, Sanjeev Raja, Ashesh Chattopadhyay, Morteza Mardani, Thorsten Kurth, David Hall, Zongyi Li, Kamyar Azizzadenesheli, et~al.
\newblock Fourcastnet: A global data-driven high-resolution weather model using adaptive fourier neural operators.
\newblock {\em arXiv preprint arXiv:2202.11214}, 2022.

\bibitem{robert1993bubble}
Andr{\'e} Robert.
\newblock Bubble convection experiments with a semi-implicit formulation of the euler equations.
\newblock {\em Journal of the Atmospheric Sciences}, 50(13):1865--1873, 1993.

\bibitem{vincent2008extracting}
Pascal Vincent, Hugo Larochelle, Yoshua Bengio, and Pierre-Antoine Manzagol.
\newblock Extracting and composing robust features with denoising autoencoders.
\newblock In {\em Proceedings of the 25th international conference on Machine learning}, pages 1096--1103, 2008.

\bibitem{Xiao_2023}
Yongqian Xiao, Zixin Tang, Xin Xu, Xinglong Zhang, and Yifei Shi.
\newblock A deep koopman operator‐based modelling approach for long‐term prediction of dynamics with pixel‐level measurements.
\newblock {\em CAAI Transactions on Intelligence Technology}, 9(1):178–196, February 2023.

\bibitem{zhu2019physics}
Yinhao Zhu, Nicholas Zabaras, Phaedon-Stelios Koutsourelakis, and Paris Perdikaris.
\newblock Physics-constrained deep learning for high-dimensional surrogate modeling and uncertainty quantification without labeled data.
\newblock {\em Journal of Computational Physics}, 394:56--81, 2019.

\end{thebibliography}

\end{document}